%% file: main.tex
\documentclass[10pt,twocolumn,letterpaper]{article}

\usepackage{iccv}              


\definecolor{iccvblue}{rgb}{0.21,0.49,0.74}
\usepackage[pagebackref,breaklinks,colorlinks,allcolors=iccvblue]{hyperref}
\usepackage{multirow}
\usepackage{float}
\usepackage{wrapfig}
\usepackage{bbding}
\usepackage[accsupp]{axessibility}

\usepackage{algorithm}
\usepackage[noend]{algpseudocode}

\def\paperID{9457} 
\def\confName{ICCV}
\def\confYear{2025}

\title{Mind the Gap: Preserving and Compensating for the Modality Gap in CLIP-Based Continual Learning}

\author{Linlan Huang$^1$, Xusheng Cao$^1$, Haori Lu$^1$, Yifan Meng$^1$, Fei Yang$^{2,1}$, Xialei Liu$^{2,1}$\thanks{Corresponding author.}\\
$^1$VCIP, CS, Nankai University \qquad $^2$NKIARI, Shenzhen Futian\\
{\tt\small \{huanglinlan, caoxusheng, luhaori\}@mail.nankai.edu.cn, \{feiyang, xialei\}@nankai.edu.cn}
}
\newcommand{\minisection}[1]{ \noindent {\bf #1}\ \ }

\begin{document}
\maketitle
\input{sec/0_abstract}    
\input{sec/1_intro}

\input{sec/2_related_work}

\input{sec/3_method}

\input{sec/4_experiments}

\input{sec/5_conclusion}
{
    \small
    \bibliographystyle{ieeenat_fullname}
    \bibliography{main}
}
\input{sec/supp}

\end{document}

%% file: sec/0_abstract.tex
\begin{abstract}
Continual learning aims to enable models to learn sequentially from continuously incoming data while retaining performance on previously learned tasks.
With the Contrastive Language-Image Pre-trained model (CLIP) exhibiting strong capabilities across various downstream tasks, there has been growing interest in leveraging CLIP for continual learning in such scenarios.
Most existing works overlook the inherent modality gap in CLIP, a key factor in its generalization and adaptability. 
In this paper, we analyze the variations in the modality gap during the fine-tuning of vision-language pre-trained models.
Our observations reveal that the modality gap effectively reflects the extent to which pre-trained knowledge is preserved.
Based on these insights, we propose a simple yet effective method, \textbf{MG-CLIP}, that improves CLIP’s performance in class-incremental learning.
Our approach leverages modality gap preservation to mitigate forgetting and modality gap compensation to enhance the capacity for new data, introducing a novel modality-gap-based perspective for continual learning. 
Extensive experiments on multiple benchmarks demonstrate that our method outperforms existing approaches without requiring additional replay data.
Our code is available at \url{https://github.com/linlany/MindtheGap}.

\end{abstract}

%% file: sec/1_intro.tex
\section{Introduction}
\label{sec:intro}

\begin{figure}
    \centering
    \includegraphics[width=0.9\linewidth]{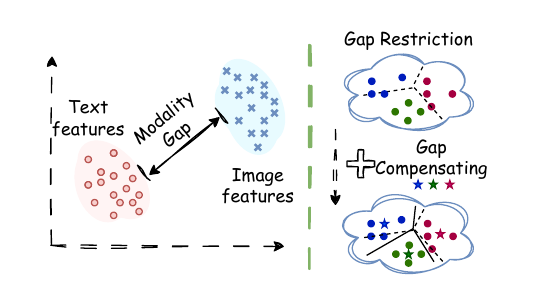}
    \caption{
    Left: In continual learning, we preserve the modality gap to retain CLIP’s original knowledge and mitigate forgetting. Right: Due to the modality gap, the text classifier's performance is restricted. We compensate for it through an intra-modal classifier by enhancing its adaptability and refining decision boundaries.
    }
    \label{fig:first}
\end{figure}

The goal of continual learning is to enable models to continuously acquire new knowledge and adapt to the ever-changing real world~\cite{van2022three_type_cl,masana2022class}. Traditional approaches to continual learning focus on training a model from scratch, aiming to reduce catastrophic forgetting of old task knowledge while maintaining plasticity to adapt to new data~\cite{li2017lwf}. This is often referred to as the stability-plasticity trade-off. However, with the recent development of large-scale pre-trained models, these models offer stronger stability and generalization capabilities for continual learning~\cite{lee2023pre_train_benefit,mehta2021pretrain_invest,wu2022classs_base_strong}. 
Among pre-trained models, visual language pre-trained models such as CLIP demonstrate excellent generalization to downstream tasks, showcasing impressive zero-shot capabilities~\cite{radford2021clip,li2024densevlm,li2024cascadeclip}. Consequently, CLIP-based continual learning (CL) has emerged as a promising new direction, attracting increasing attention from researchers~\cite{thengane2022continualclip,jhaclap4clip,yu2024moeclip,huang2024rapf}.

Existing CLIP-based continual learning methods can be broadly categorized into two approaches: one fine-tunes the backbone to modify feature representations~\cite{yu2024moeclip,marczak2024magmax}, while the other freezes the backbone and introduces learnable modules for continual learning~\cite{huang2024rapf,jhaclap4clip,zhou2023proof}.
They often treat CLIP as a feature extractor, approximating it to a text-enhanced visual model~\cite{marczak2024magmax}. 
Based on this view, they focus on better feature fusion for improved performance~\cite{zhou2023proof,jhaclap4clip} or leveraging textual information to guide visual feature adaptation~\cite{huang2024rapf}.
However, these methods often overlook CLIP’s unique cross-modal properties, the modality gap, distinguishing it from unimodal systems.
The modality gap is observed in previous work~\cite{liang2022mind_the_gap}.
As shown in Fig.~\ref{fig:first} (left), the modality gap is that the intra-modal feature distance is small while the inter-modal distance is large.
The feature distributions of the two modalities are represented in two distinct cones, with an inherent spacing between them. 
To fully exploit CLIP’s cross-modal properties, it is crucial to explore strategies that preserve the inherent modality gap while addressing its limitations in continual learning. 

In this paper, we investigate CLIP-based class-incremental learning from the perspective of the modality gap.
In existing studies on multimodal model training and adaptation of downstream tasks, the modality gap is often seen as a source of suboptimal performance. Researchers focus on reducing the modality gap in pre-trained multimodal models~\cite{eslami2024mitigate_gap_cliptrain,hu2024reclip,mistretta2025cross_the_gap}. 
However, in the context of continual learning with CLIP, our objective is to retain the strong generalization capability of the model while learning new data.
Thus, how to handle the modality gap in continual learning remains an open question. Our work is based on the assumption that this modality gap reflects the intrinsic knowledge of the pre-trained model.

Downstream datasets are typically much smaller than pretraining datasets, making them insufficient to fully train the model. As a result, modifying the fundamental properties of a pre-trained model, such as the modality gap, based on the downstream data may disrupt the pre-trained knowledge. It may potentially compromise the performance of continual learning.
During training on the current task, the cross-entropy optimization objective tends to expand this gap. This phenomenon intensifies as incremental learning data continue to arrive. 
Furthermore, directly aligning the two modalities for downstream tasks will also significantly alter the pre-trained knowledge. 
Therefore, we keep the modality gap relatively stable to maintain the stability of the model.
We analyze the changes in modality gap during training and propose a modality gap-aware adjustment strategy.
By tracking variations in the modality gap, we regulate the training to maintain a stable modality gap, thereby preserving pre-trained knowledge and mitigating forgetting.

Additionally, we analyze the impact of preserving the modality gap. 
As shown in Fig.~\ref{fig:first} (right), when text features are used as classifiers, the modality gap may restrict the model's ability to learn new data in continual learning.
This limits the adaptability of the model, reducing the plasticity. To compensate modality gap, we propose to build a classifier in the visual space, where the modality gap does not pose a restriction. 
By integrating its output with that of the text classifier, we compensate for the modality gap and improve the learning capacity of CLIP.

The main contributions of this paper are as follows:

\begin{itemize}
    \item We propose a method to preserve the modality gap that maintains pre-trained knowledge and mitigates forgetting in continual learning.
    \item We compensate for the limitations of the modality gap by introducing a complementary mechanism, enhancing model plasticity.
    \item Our approach achieves state-of-the-art results across multiple datasets without replay or complex structures.
\end{itemize}

%% file: sec/2_related_work.tex
\section{Related Work}
\label{sec:related_work}
\subsection{Class-Incremental Learning}
Class-incremental learning methods are generally grouped into three types~\cite{de2021continual_type}.
Regularization methods limit changes to the model. Some methods use distillation to reduce the deviation of model features or penalize changes in model parameters~\cite{kirkpatrick2017ewc,li2017lwf,zhang2020DMC,choi2021dual_teacher_datafree,zhou2021coil}. 
Dynamic network approaches allow the model structure to evolve as new tasks are introduced~\cite{douillard2022dytox,wang2022foster,wu2022classs_base_trong,yan2021der}. Replay-based methods retain original samples or relevant information from them. When learning a new task, these methods replay original samples~\cite{rebuffi2017icarl,lopez2017gem,sun2023Second_Order_select} or recover old samples from the retained information~\cite{huang2024rapf,zhang2023slca,zhai2023mae,wang2021memory_compre}. The latter often involves techniques such as memory compression or feature replay. Dynamic networks or feature replay methods tend to increase both parameter count and memory storage requirements. With the widespread use of pre-trained models, incremental learning methods for visual pre-trained models have also emerged. Parameter-efficient fine-tuning approaches gradually expand a small number of parameters as tasks increase~\cite{wang2022l2p,wang2022dualprompt,smith2023coda}. Aper~\cite{zhou2024revisiting_adam_aper}  maintains stability by training the model only on the first task. SLCA~\cite{zhang2023slca} uses a small learning rate fine-tuning backbone for continual learning. 
They show that preserving the stability of pre-trained visual models enhances generalization and benefits class-incremental learning on downstream tasks.

\subsection{Class-Incremental Learning with CLIP}
CLIP performs remarkably well in class-incremental learning tasks with downstream data~\cite{thengane2022continualclip}. Continual learning based on CLIP has gained increasing attention.
Some approaches add learnable modules to the original CLIP features to modify them for better adaptation to new tasks.
PROOF~\cite{zhou2023proof} and CLAP~\cite{jhaclap4clip} add learnable modules to the CLIP output features to facilitate cross-modal interaction. RAPF~\cite{huang2024rapf} introduces a linear layer after the CLIP visual encoder for downstream adaptation and uses textual modality information to guide feature replay. 
Other methods fine-tune CLIP to alter its output features. 
ZSCL~\cite{zheng2023ZSCL} distills additional datasets during training. MOE4CL~\cite{yu2024moeclip} fine-tunes CLIP with a mixture of experts, introducing a selection mechanism to selectively use the original CLIP model. Magmax~\cite{marczak2024magmax} sequentially fine-tunes the entire CLIP model, using a task vector algorithm to merge models at inference time. LGVLM~\cite{zhang2024LGVLM} trains separate LoRA modules for each task.
These methods primarily focus on leveraging the prior knowledge of natural language to assist continual learning, often neglecting the preservation of zero-shot capabilities while emphasizing performance on sequential tasks. In contrast, we focus on retaining the model's capabilities while enhancing its continual learning ability, starting from the inherent property of the CLIP.

\subsection{Modality Gap}
Liang \etal~\cite{liang2022mind_the_gap} demonstrated a phenomenon in pre-trained vision language models called the modality gap. 
In contrastive vision-language models, the modality gap is reflected in the fact that text and image features are located in two separate narrow cones. 
The features of different modalities exhibit distinct separation, whereas the features within the same modality tend to cluster more closely.
Existing studies have explored the impact of the modality gap on tasks such as domain adaptation~\cite{hu2024reclip}, few-shot classification~\cite{yi2024leveraging_gap}, model pre-training~\cite{fahim2024address_gap} and retrieval task~\cite{mistretta2025cross_the_gap}. Reducing the modality gap can improve the model performance in these tasks. However, in continual learning tasks, the stability of pre-trained knowledge is more crucial for downstream tasks over time. Therefore, we propose using the modality gap as an indicator of the changes in the pre-trained model's feature space, with the goal of preserving the modality gap in continual tasks.

%% file: sec/3_method.tex
\section{Preliminaries}

\paragraph{Class incremental learning definition.}
We consider a class-incremental learning setup based on a pre-trained CLIP model $ M $. 
The objective is to sequentially train the model on a series of classification tasks. 
Each task $ t $ consists of a set of categories $ C_t $, and the categories between tasks do not overlap, $\forall i \neq j, C_i\cap C_j = \emptyset$. 
During the training of task $ t $, the model has no access to information related to the previous $ t-1 $ tasks. 
After training in the $ t $-th task, the model $ M_t $ is required to correctly classify all previously learned categories, $ C_1 \cup C_2 \cup \dots \cup C_t $, without access to task identifiers.

\paragraph{Modality gap measure.}
In the classification task, given $ N $  images and $ K $ class name of text, we measure the inter-modality similarity as the average cosine similarity between all image and text features: $\frac{1}{N} \sum_{i=1}^{N} \frac{1}{K} \sum_{j=1}^{K} \cos(\mathbf{x}_i, \mathbf{t}_j)$, where $\mathbf{x}_i$ is an image feature and $\mathbf{t}_j$ is a text feature. 
This captures the overall similarity between the image and text modalities, reflecting the modality gap in the feature space. 

To analyze the impact of image-to-text similarity in more fine-grained, we define the average similarity of positive image-text pairs as:
\begin{equation}
    pos= \frac{1}{N}\sum_i^N\cos ( \mathbf{x}_i, \mathbf{t}_{yi}),
    \label{eq:pos}
\end{equation}
where $\mathbf{t}_{yi}$ denotes the corresponding class text feature for the image $\mathbf{x}_i$. Similarly, we measure the average similarity of negative image-text pairs as:
\begin{equation}
    neg = \frac{1}{N} \sum_{i=1}^{N} \frac{1}{K-1} \sum_{j=1}^{K-1} \cos(\mathbf{x}_i, \mathbf{t}_j^{neg}),
    \label{eq:neg}
\end{equation}
where $\mathbf{t}_j^{neg}$ represents the text features of non-matching classes for the image $\mathbf{x}_i$.

\section{Method}

Our method approaches CLIP-based continual learning from the perspective of the modality gap, recognizing its crucial role in preserving CLIP’s generalization capabilities during continual learning. 
We identify that uncontrolled training can disrupt this inherent modality gap, leading to suboptimal performance in downstream incremental learning. 
To address this, we propose a two-stage strategy: (1) Preserving the modality gap for model stability. We regulate the training process to ensure a stable modality gap across tasks; (2) Compensating the modality gap for model plasticity. The constraint for the modality gap may limit the model’s adaptability, we introduce a complementary classifier that enhances task-specific performance without altering the preserved modality gap.
During inference, we integrate both outputs to balance stability and task plasticity. Next, we discuss the motivation (Sec.~\ref{motivation}) and details of our method (Sec.~\ref{preserve_gap} and Sec.~\ref{compensate_gap}). The pseudo-code can be found in Sec.5 of the supplementary material.

\subsection{Efffect of Modality Gap in Continual Learning}
\label{motivation}
In this section, we begin by analyzing the evolution of the modality gap during continual learning.

\paragraph{Cross-entropy loss expands the modality gap.}

The optimization objective of cross-entropy is not consistent with the modality gap of the original CLIP.
When optimized via cross-entropy loss, the training objective aligns image-text pairs to a cosine similarity of 1 for matches and -1 for non-matches. 
However, this optimization contradicts the moderate similarity distribution of the original CLIP model, which ranges from approximately 0 to 0.3 reflecting the inherent modality gap. 
It causes the modality gap of the trained model to expand.
Furthermore, downstream classification tasks introduce a structural imbalance absent in CLIP pre-training. Unlike symmetric text-image pairs in pre-training, classification tasks pair multiple images of the same class with a single text embedding. In a $C$-class dataset, each text forms positive pairs with only $\frac{1}{C}$ of the images while being repelled from the remaining $\frac{C-1}{C}$ negative pairs. The imbalance leads to the optimization process dominated by repelling non-matching samples, further expanding the modality gap.

\begin{figure}[t]
    \centering
    \begin{subfigure}[t]{0.5\linewidth}
        \includegraphics[width=\linewidth]{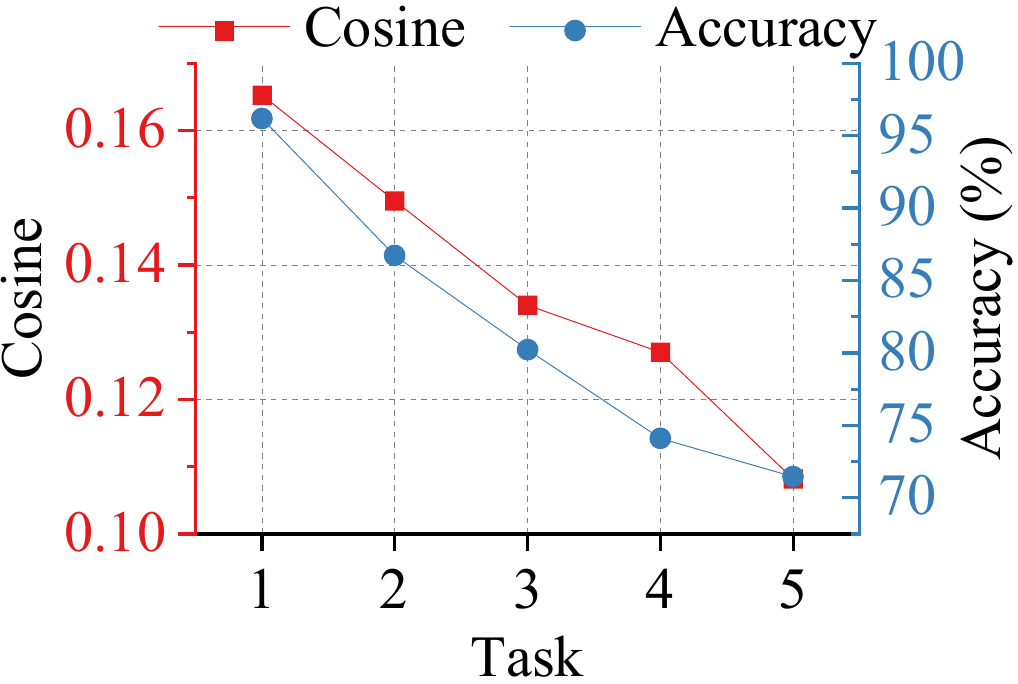}
        \caption{Naive fine-tuning reduces cosine similarity, widens modality gap, and produces significant forgetting.}
        \label{fig:big_gap}
    \end{subfigure}
    \begin{subfigure}[t]{0.49\linewidth}
        \includegraphics[width=\linewidth]{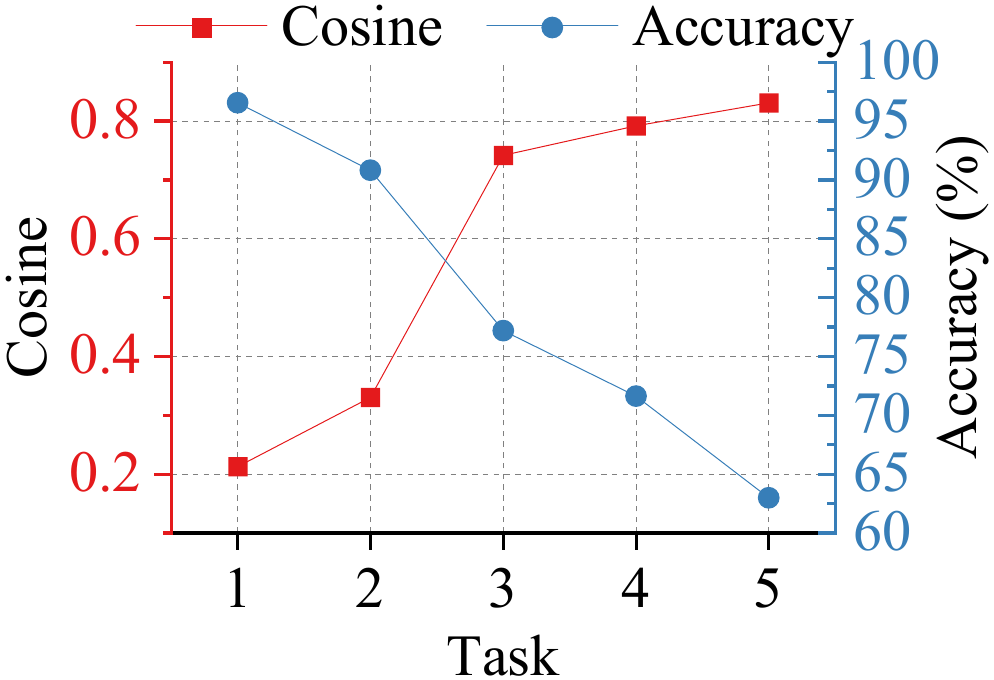}
        \caption{The alignment loss increases cosine similarity and reduces modality gap, but it disrupts pre-trained knowledge, leading to forgetting.}
        \label{fig:small_gap}
    \end{subfigure}
    \caption{Average cosine similarity between text and image features, along with accuracy on ImageNet-R during CL.}
    \label{fig:gap_change}
\end{figure}

The modality gap implicitly reflects pre-trained knowledge, and expanding it may result in forgetting previous knowledge.
As shown in Fig.~\ref{fig:big_gap}, experimental results confirm this effect: the cosine similarity between image and text representations steadily declines as tasks progress, demonstrating the growing modality gap. This coincides with a continuous drop in accuracy, highlighting its adverse impact on CLIP’s pre-trained knowledge and overall performance in class-incremental learning.

\paragraph{Direct alignment loss reduces modality gap but disrupts pre-trained knowledge.}
As shown in Fig.~\ref{fig:small_gap}, introducing a loss of alignment~\cite{fahim2024address_gap} in downstream tasks can reduce the modality gap by minimizing the Euclidean distance between matched image and text features.  
The key issue lies in the mismatch of the subspace: downstream datasets span a limited region of the original CLIP feature space, which makes the direct alignment risk of overspecialization.
Directly aligning the two modalities can significantly alter pre-trained representations, leading to forgetting. 
\paragraph{Sustaining CLIP's lifelong learning capacity.}
Previous analyses have revealed a phenomenon: naive fine-tuning tends to expand the modality gap, while direct alignment reduces it. 
As class-incremental learning progresses, this problem becomes increasingly severe, gradually eroding CLIP’s pretraining capabilities. Since CLIP’s pretraining knowledge plays a crucial role in providing stability for downstream tasks, we need to maintain a relatively stable modality gap. 
Ensuring this stability is essential for allowing CLIP to incorporate new knowledge without compromising its fundamental vision-language correspondence.

\begin{figure}
    \centering
    \begin{subfigure}[t]{0.49\linewidth}
        \includegraphics[width=\linewidth]{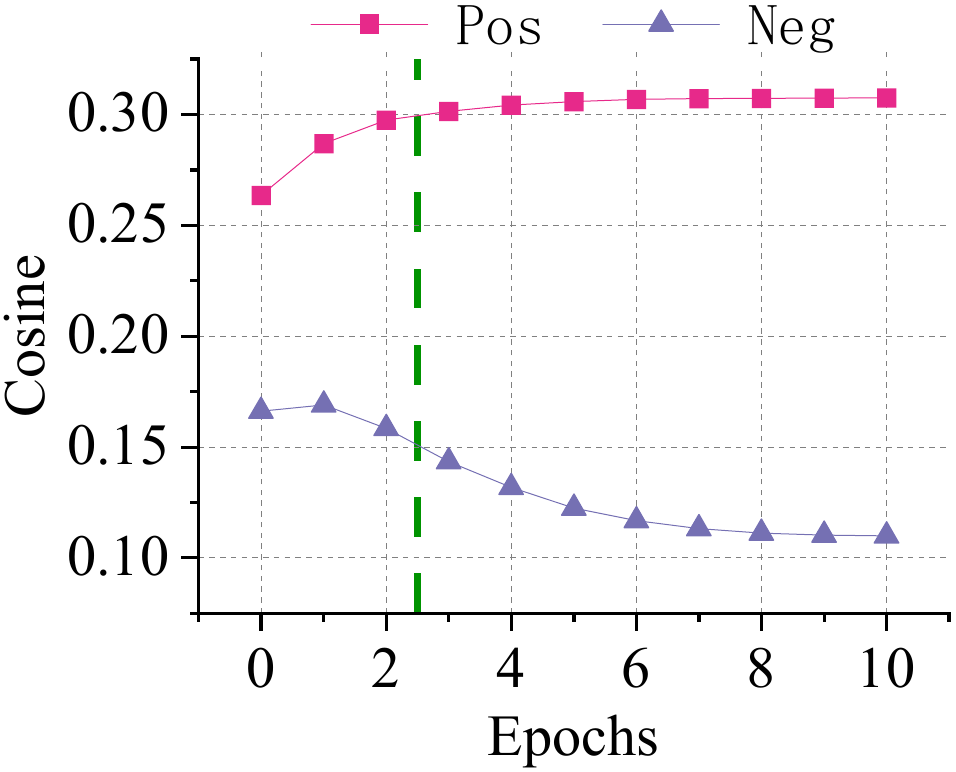}
        \caption{ImageNet-R}
        \label{fig:change_intask_imgr}
    \end{subfigure}
    \begin{subfigure}[t]{0.49\linewidth}
        \includegraphics[width=\linewidth]{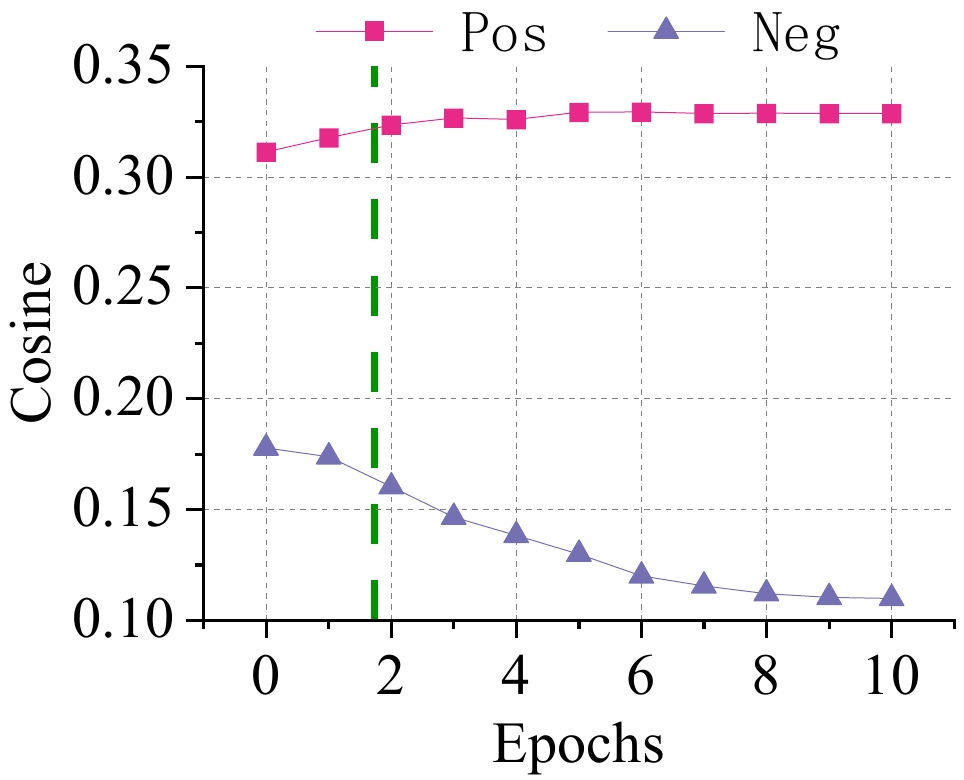}
        \caption{ImageNet-100}
        \label{fig:change_intask_img100}

    \end{subfigure}

    \caption{The variation in the mean of positive and negative cosine within a task during training. The cosine change pattern before and after the green dotted line is different.}
    \label{fig:intask_gap_change}
\end{figure}

\subsection{Adaptive Modality Gap Preservation}
In this section, we characterize the asymmetric evolution of the modality gap in continual learning and propose an adaptive preservation strategy to enhance model stability.

\begin{figure*}
    \centering
    \includegraphics[width=0.92\linewidth]{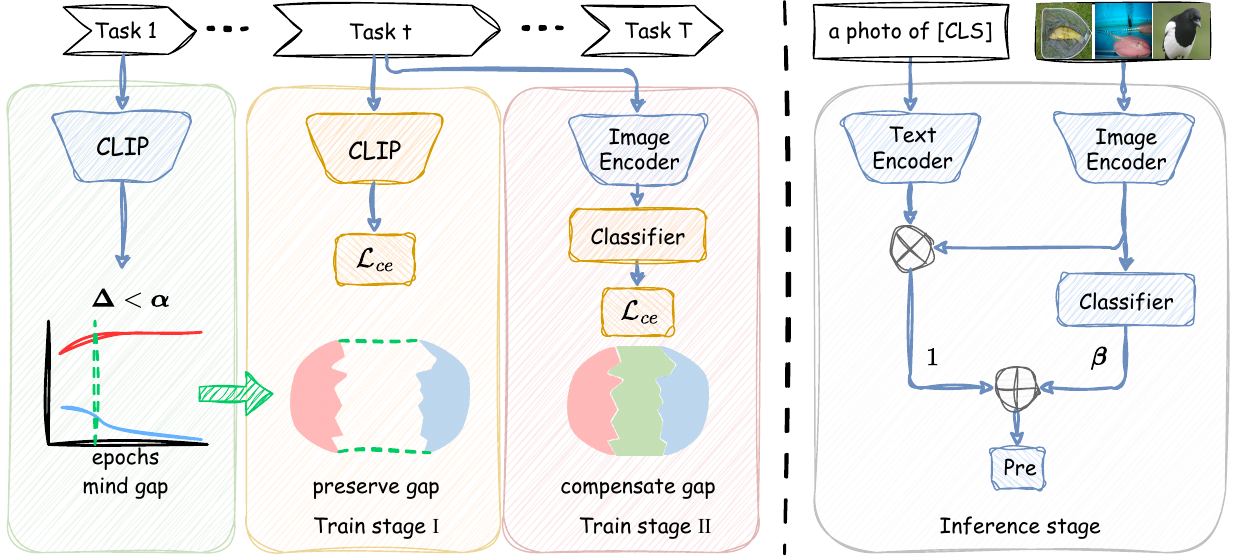}
    \caption{Our method tackles CLIP-based continual learning through dual mechanisms: (1) {\bf Modality Gap Preservation} stops training when modality gap deviation exceeds a stability threshold, preventing cross-modal knowledge distortion; (2) {\bf Gap Compensation} trains a visual-space classifier on frozen features to enhance task-specific plasticity while respecting preserved gap. In inference stage, we combine two different classifier subspaces for prediction.}
    \label{fig:pipeline}
\end{figure*}

\paragraph{Asymmetry phenomenon of modality gap change during training.}
\label{preserve_gap}
As shown in Fig.~\ref{fig:intask_gap_change}, we observe an asymmetric change in cosine similarity between text and image features during single task training. 
The `pos' and `neg' use Eq.~\ref{eq:pos} and Eq.~\ref{eq:neg} to calculate.
Before training, \ie, epoch 0, the `pos' is already larger than the `neg', indicating the zero-shot capability of the model. Even the similarity between an image and its corresponding class text remains relatively low, which is a reflection of the modality gap.
During the early phase of training, the output changes are primarily due to an increase in the similarity of positive image-text pairs, indicating the model has learned new knowledge and becomes more confident about the true class. This corresponds to the part before the green dashed line in Fig.~\ref{fig:intask_gap_change}.
As long as the positive output remains larger than the negative outputs, the model can still make correct predictions even if the loss exists.
However, in the later stages of training, the positive output stabilizes and the negative outputs begin to decrease. This indicates that the distance between the image and most other text features increases. 
Therefore, an optimal training stage can retain its learned knowledge while keeping the inter-modal distance relatively stable.

\paragraph{Adaptive training for preserving modality gap.}
Based on these observations, we propose determining the number of training epochs by monitoring the mean negative outputs. 
As shown in Fig.~\ref{fig:pipeline}, using the first task data of the dataset, we estimate the required number of epochs for subsequent tasks.
Specifically, we first compute the mean of the negative outputs using Eq.~\ref{eq:neg}, denoted as $ neg^0 $, using the original CLIP model for the first task. Then, we train the model using LoRA, and after the $e$ epoch, we compute the $ neg^e $ of the negative outputs for all data. We calculate the relative difference between the current and the original as follows:
\begin{equation}
    \Delta = \frac{\left| neg^e - neg^0 \right|}{neg^0}.
\label{eq:Relative_diff}
\end{equation}
When the difference $\Delta$ exceeds a predefined threshold $\boldsymbol{\alpha}$, we record the last epoch $ e $ where $\Delta$ was still below $\alpha$. Finally, for all tasks in the downstream dataset, we train the model for the $ \max(e, 1) $ epochs.

\subsection{Intra-modal Compensation for Modality Gap}
\label{compensate_gap}
In this section, we explain how the modality gap limits the model's capability, as it must be maintained as discussed in the previous section. Consequently, we construct an intra-modal classifier to compensate for the limitations.

\paragraph{Modal gap restriction on text classifier capability.}

To achieve minimal cross-entropy loss, an optimal classifier must exist within the image feature space. Specifically, for any classifier that minimizes classification error, there exists an equivalent form $\mathbf{W}_{\text{opt}}$ entirely contained in the span of image features. 
This follows from the decomposition of any classifier into components parallel and orthogonal to the image space, where the orthogonal component does not contribute to classification. 
A detailed proof of this claim is provided in the supplementary material 2.1.

Given this, we analyze the modality gap's impact on text classifiers. The text feature matrix $ \mathbf{T} $ decomposes as:  
$
\mathbf{T} = \mathbf{T}_{\parallel} + \mathbf{T}_{\perp}
$
where $ \mathbf{T}_{\parallel} $ lies in the subspace spanned by visual features $ \mathbf{X} $, and $ \mathbf{T}_{\perp} $ is orthogonal to it. Since text features generally do not fully span the image feature space, the best achievable text classifier is restricted to a low-rank subspace, leading to misalignment error. The lower bound of the distance from $ \mathbf{T}_{\parallel} $ to the optimal image-space classifier is determined by the singular values $ s^2 $ outside the text feature subspace:
\begin{equation}
    \|\mathbf{T}_{\parallel} - \mathbf{W}_{\text{opt}}\|_F^2 
    \geq \sum\nolimits_{i=r+1}^{r'} s_i^2,
\end{equation}
where $r'$ is the rank of $\mathbf{W}_{\text{opt}}$ and $r$ is the rank of $\mathbf{T}_{\parallel}$. 
The formal derivation of this bound is provided in the supplementary material 2.2.

This result reveals an inherent limitation: unless the text feature space has sufficient capacity to represent the optimal classifier, perfect alignment is unattainable. 
Due to the modality gap, text classifiers often operate in a lower-rank subspace, restricting their classification effectiveness.

\paragraph{Modal gap compensation via intra-modal classifier.}
To compensate the modality gap, we introduce an auxiliary classifier in the visual space. 
The fine-tuned CLIP model $ f_{\text{clip}}(\cdot) $ and the classifier weights for old classes are frozen.
For classes introduced in the current task, we initialize their classifier weights within the cosine classifier $\mathbf{W}_v$ using their class prototypes and train them using image features without text.
Since the gradients of the classifier remain in the input space, which is the visual space, this ensures that the classifier operates within the visual subspace.

As shown in Fig.~\ref{fig:pipeline}, in the model inference stage, we combine the predictions from both the text and visual classifiers. The final predict score is calculated as:
\begin{equation}
    pre(\mathbf{x}) = f_{clip}(\mathbf{x},\mathbf{t}) + \boldsymbol{\beta}\cdot softmax(\mathbf{W}_v^\top\mathbf{x}),
    \label{eq:output}
\end{equation}
where $\boldsymbol{\beta}$ is a constant hyperparameter.

%% file: sec/4_experiments.tex
\section{Experiments}
\subsection{Experimental Setup}

\begin{table*}[]
    \centering

    \begin{tabular}{cccccccccccc}
    \toprule
    \multirow{2}{*}{Method}
    & \multirow{2}{*}{Exemplar}
    &\multicolumn{2}{c}{CIFAR-100} 
    &\multicolumn{2}{c}{ImageNet-R} 
    &\multicolumn{2}{c}{ImageNet100} 
    &\multicolumn{2}{c}{ImageNet-1K} 
    &\multicolumn{2}{c}{VTAB}\\
    &&Avg &Last &Avg &Last &Avg &Last &Avg &Last &Avg &Last\\
    \midrule
    PROOF~\cite{zhou2023proof}&\multirow{2}{*}{DR} &84.88  &76.29  &82.83 &77.05   &84.71 &72.48   &76.23&65.26    &89.09&83.97\\
    CLAP~\cite{jhaclap4clip}&                               &86.13  &78.21  &85.77 &79.98   &87.76 &79.16   &81.72&73.19              &91.37&89.67\\
    \midrule
    SLCA~\cite{zhang2023slca}&\multirow{2}{*}{FR}&80.53 &67.58  &75.92 &70.37   &78.63 &59.92   &79.10 &68.27   &84.25&82.54\\
    RAPF~\cite{huang2024rapf}&                              &86.19  &79.04  &85.58 &80.28   &87.51 &80.23   &81.73 &72.58   &90.88&82.31\\
    \midrule
    L2P++~\cite{wang2022l2p}&\multirow{8}{*}{NR}     &81.90  &73.08  &81.67 &75.98   &80.51 &67.22   &79.30 &69.60   &63.23&38.37\\
    DualPrompt~\cite{wang2022dualprompt}&                   &81.45  &72.51  &82.01 &75.77   &80.65 &67.38   &79.39 &69.79   &61.89&37.58\\
    CODA~\cite{smith2023coda}&                            &76.98  &62.25  &78.00 &67.52   &64.13 &34.76   &76.99 &66.96   &62.51&38.25\\
    Continual-CLIP~\cite{thengane2022continualclip}&        &75.15  &66.68  &79.12 &72.00   &84.98 &75.40   &72.96 &64.44   &53.64&31.50\\
    Aper-Adapter~\cite{zhou2024revisiting_adam_aper}&       &75.76  &65.50  &78.65 &71.35   &85.84 &76.40   &76.60 &68.74   &80.75&71.21\\
    MOE4CL~\cite{yu2024moeclip}&                            &85.36  &78.37  &85.28 &80.77   &86.39 &76.66   &81.29 &72.73   &68.49&61.70\\
    CLAP*~\cite{jhaclap4clip}&                              &74.19  &63.45  &81.22 &75.80   &81.07 &72.00   &75.85&67.36             &82.11&80.11\\
    MagMax~\cite{marczak2024magmax}&                        &85.63  &79.00  &87.13 &80.85   &86.33 &75.92   &80.74 &71.31   &64.63&53.90\\
    \midrule
    MG-CLIP (Ours)&NR                                          &87.00  &80.57  &87.58 &82.67   &87.31&78.38   &81.88 &73.68   &94.67&91.53\\
    \bottomrule
    \end{tabular}

    \caption{Comparison of performance with different methods. DR denotes using real data for replay, FR denotes generating old class features for replay, and NR denotes non-replay. The results are mainly obtained from references~\cite{huang2024rapf,jhaclap4clip} and reproduced using their publicly available code. The performance of ours is the average over three different class orders. Except for VTAB, which is divided into 5 tasks, the other datasets are divided into 10 tasks. CLAP* represents the replay-free version provided by the paper of CLAP.}
    \label{tab:main_exp}
\end{table*}

\minisection{Datasets.}
We evaluate our method through continual learning tasks on five benchmark datasets: CIFAR-100~\cite{krizhevsky2009cifar100}, ImageNet-R~\cite{hendrycks2021imagenet-r}, ImageNet-100~\cite{deng2009imagenet}, ImageNet-1K~\cite{deng2009imagenet}, and VTAB~\cite{zhai2019vtab}.
All of these datasets except the VTAB are evenly divided into 10 sequential tasks. 
Following the previous work~\cite{zhou2024revisiting_adam_aper}, we extracted a subset from VTAB, including 5 tasks, each with ten categories. For more experimental details, please refer to Sec.4 of the supplementary material.

\minisection{Competing methods.}
Our experiments compare two types of methods: (1) Vision-only approaches that include L2P++~\cite{wang2022l2p}, DualPrompt~\cite{wang2022dualprompt}, CODA~\cite{smith2023coda}, SLCA~\cite{zhang2023slca}, and Aper-Adapter~\cite{zhou2024revisiting_adam_aper}; (2) CLIP-based methods that include PROOF~\cite{zhou2023proof}, CLAP~\cite{jhaclap4clip}, RAPF~\cite{huang2024rapf}, MOE4CL~\cite{yu2024moeclip}, MagMax~\cite{marczak2024magmax} and the zero-shot CLIP baseline Continual-CLIP~\cite{thengane2022continualclip}. All methods adopt the ViT-B/16 weights of OpenAI~\cite{radford2021clip} by default. While original MagMax implementations employ enhanced augmentations and optimized text templates, we follow previous works~\cite{yu2024moeclip,huang2024rapf,yu2024moeclip} and standardize evaluations using basic image augmentations and a fixed prompt template for fair comparison.

\minisection{Evaluation metrics.}
The average accuracy on the test data after training the t-th task, considering the first through $t$ tasks, is represented as $A_t$. 
`Avg' is the average of the accuracies of all tasks, \ie, $\frac{1}{t}\sum_{i=1}^tA_i$. `Last' indicates the average accuracy after the final task $T$, \ie, $A_{T}$.

\minisection{Implementation detail.}
We develope our method using PyTorch on an A40 GPU. Unless otherwise specified, we use OpenAI's pre-trained CLIP model, specifically the ViT-B/16 version. Results for another version of CLIP can be found in the supplementary materials.
During the training of the backbone, we apply LoRA~\cite{hu2022lora} for model adaptation, with the default rank set to 8. 
Since the focus of our work is not on the fine-tuning process itself, we simplify the implementation by applying LoRA only to the key and value parts of the attention module. Different fine-tuning model implementations are left for future work.
We use the Adam optimizer with a cosine learning rate scheduler with an initial learning rate of 0.001. The number of epochs is determined by the first part of the method, where the threshold $\boldsymbol{\alpha}$ is set to 10\%. 
In the training stage for the image-space classifier, we train for 3 epochs, with an initial learning rate of 0.0005. The classifier uses a cosine classifier. 
During inference, the hyperparameter $\boldsymbol{\beta}$ to integrate the two classification results is set by default to 4. Hyperparameter effects are described in the supplementary material.

\subsection{Comparison Results}

Table~\ref{tab:main_exp} presents a comparison of our approach with other methods. In most datasets, our method outperforms all others, including those relying on replay. Specifically, on CIFAR-100, our method surpasses all competitors in Last accuracy by at least 1.53\%. 
On ImageNet-R, we achieve an improvement of at least 1.82\% in Last accuracy. 
On ImageNet-100, while our approach performs slightly worse than RAPF and CLAP, both of which use replay, we do not rely on replay and still outperform all non-replay methods by at least 1.72\% in Last accuracy. 
Additionally, on ImageNet-100, CLAP* performs significantly worse than CLAP, indicating that much of CLAP’s performance gain stems from data replay. 
On the larger-scale ImageNet-1K dataset, our method achieves comparable or even slightly better results than the best-performing alternatives.

The VTAB dataset poses a significant challenge to CLIP, as evidenced by the zeroshot accuracy, \ie, Continual\_CLIP, which is only 31.5\%. 
On this challenging benchmark, most methods suffer substantial performance drops. 
Our method outperforms the best replay-based CLAP approach by 1.86\% in Last accuracy, significantly outperforming its non-replay version. 
It shows a clear advantages of our method in this challenging cross-domain scenario.

\subsection{Comparison of the Zero-shot Capability}

Continual learning aims to allow models to incrementally acquire new knowledge while retaining existing capabilities. CLIP models, compared to traditional pretrained vision models, demonstrate superior zero-shot generalization. Therefore, continual learning with CLIP should not only minimize forgetting on new downstream tasks but also preserve its original zero-shot abilities. This ensures CLIP based continual learning is more than just a method for task-specific initialization that leads to overfitting, a limitation of previous works in this area. These works often follow traditional evaluation protocols, neglecting the preservation of the model’s intrinsic abilities. We propose evaluating zero-shot generalization of CLIP model on independent benchmarks after fine-tuning to address this issue.

As shown in Tab.~\ref{tab:zero_shot}, we assess zero-shot performance on three standard datasets following continual learning on CIFAR-100. Our approach yields slight improvements over the original CLIP on Food101~\cite{bossard2014food} and Oxford Pets~\cite{parkhi2012pets}, which are clearly distinct from CIFAR-100, while all baselines show performance degradation. On the more similar ImageNet-1K, three methods outperform the original CLIP, with our method achieving the most substantial improvement (a 2.85\% gain). These results suggest that our approach successfully preserves pre-trained knowledge while effectively incorporating new information.

Notably, replay-based methods (PROOF, RAPF, and CLAP) experience more significant performance declines in zero-shot tasks compared to non-replay methods. This indicates that replay mechanisms may lead to overfitting to downstream tasks, whereas non-replay methods must preserve original representations to prevent forgetting. This further highlights the potential negative impact of replay strategies on pre-trained models.

\begin{table}[]
    \centering
    \begin{tabular}{ccccc}
    \toprule
    &Food101&Pets&ImageNet-1K&Avg\\
    \midrule
    CLIP&85.14	&87.6	&64.44	&79.06\\
    \midrule
    PROOF&9.75	&22.81	&11.18	&14.58\\
    RAPF&17.18	&28.56	&15.2	&20.31\\
    CLAP&80.82	&74.68	&56.04	&70.51\\
    MOE4CL&82.85	&84.06	&66.02	&77.64\\
    MagMax&81.67	&85.96	&66.44	&78.02\\
    \midrule
    Ours&85.70	&88.17	&67.29	&80.39\\
    \bottomrule
    \end{tabular}
    \caption{Zero-shot performance of the model on different downstream datasets after completing all class-incremental learning tasks on CIFAR-100. CLIP refers to the original CLIP pre-trained model without any downstream task-specific training.}
    \label{tab:zero_shot}
\end{table}

\begin{figure}
    \centering
    \includegraphics[width=0.78\linewidth]{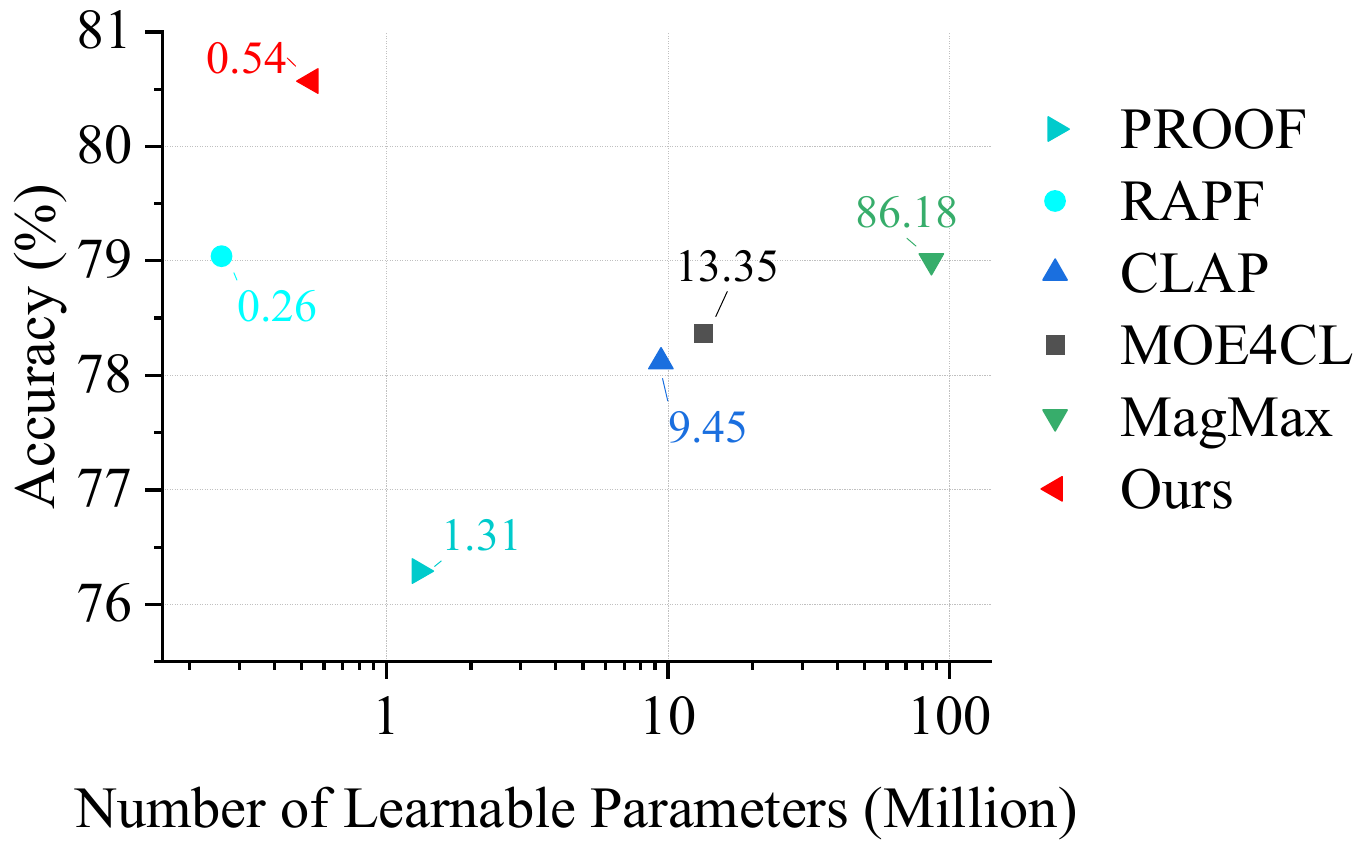}
    \caption{Comparison of accuracy and learnable parameters of different methods.}
    \label{fig:parameters}
\end{figure}
\subsection{Training Cost Analysis.}
Using CIFAR-100 as an example, we analyze the parameter overhead of different methods by comparing the additional learnable parameters introduced by our approach and other baselines, as illustrated in Fig.~\ref{fig:parameters}.
Our method introduces only 0.54M additional trainable parameters. 
Although the RAPF method requires fewer learnable parameters, it still necessitates storing a covariance matrix for each class, leading to additional storage consumption proportional to the square of the feature dimension, \ie, $nd^2$. For instance, with 100 classes, this results in an extra storage overhead of over 26M parameters. 
In contrast, our method avoids replay entirely, maintaining minimal storage consumption.

In the training process, the only extra cost in our method occurs after each epoch of the first task, where we perform an additional forward pass on the data of task to evaluate change of the modality gap. However, this step only needs to be done for the first task, does not involve backpropagation, and is minimal compared to the training cost.

\subsection{Ablation Study}

\minisection{Module ablation.}
Table~\ref{tab:module_ablation} presents the ablation study of our proposed modules. The baseline setup involves simple fine-tuning of the CLIP model for each task. Following the settings of previous work, we fine-tuned each task for 10 epochs~\cite{marczak2024magmax} and used the widely adopted cross-entropy loss with the old-class output mask~\cite{yu2024moeclip}. MGC represents Modality Gap Compensation, while MGP stands for Modality Gap Preservation.
As shown in Tab.~\ref{tab:module_ablation}, both components of our method, when used individually, lead to performance improvements, with the modality gap preservation(MGP) providing a more significant boost. The best performance is achieved when both components are used together.

When using MGC alone in the baseline setup, the improvement in last' accuracy is 1.84\%, which is larger than the improvement (1.52\%) when adding MGC to MGP. A similar trend is observed for `Avg' accuracy.
This suggests that simple fine-tuning in the baseline setup has expanded the modality gap, and therefore the role of MGC in compensating for the gap becomes more pronounced.

\begin{table}[]
    \centering
    \begin{tabular}{cccc}
    \toprule
    MGP & MGC & Avg & Last \\
    \midrule
    \XSolidBrush&\XSolidBrush&84.30&72.74\\
    \XSolidBrush&\Checkmark&85.10&74.58\\
    \Checkmark&\XSolidBrush&86.73&76.86\\
    \midrule
    \Checkmark&\Checkmark&\textbf{87.31}&\textbf{78.38}\\
    \bottomrule
    \end{tabular}
    \caption{Ablation study of the modules on ImageNet-100. MGP refers to our adaptive modality gap preservation, and MGC denotes the intra-modal Compensation for Modality Gap.}
    \label{tab:module_ablation}
\end{table}

\minisection{Image space and classifier space.}
We investigate the relationships among the linear subspaces spanned by the image features, the text-based classifier, and the learned modality-gap-compensating classifier.
To quantify their differences, we apply matrix decomposition to extract the orthonormal bases of each space. For detailed calculations, refer to the supplementary materials Sec.3.
$\mathbf{B}_i$: Orthonormal basis of image feature space;
$\mathbf{B}_t$: Orthonormal basis of text classifier;  
$\mathbf{B}_{vc}$: Orthonormal basis of the modality-gap-compensating classifier;  
$\mathbf{B}_{t+vc}$: Orthonormal basis of the combined space spanned by both classifiers.  
To measure how well the text classifier space $\mathbf{B}_t$ covers the image feature space $\mathbf{B}_i$, we compute the mean norm of the orthogonal component of $\mathbf{B}_i$ with respect to $\mathbf{B}_t$:  
\begin{equation}
    d(\mathbf{B}_i, \mathbf{B}_t) = \frac{1}{|\mathbf{B}_i|} \sum_{\mathbf{x} \in \mathbf{B}_i} \| \mathbf{x} - \mathbf{B}_t\mathbf{B}_t^\top\mathbf{x} \|.
    \label{eq:res_spcae}
\end{equation}
This metric reflects how much of $ \mathbf{B}_i $ lies outside $ \mathbf{B}_t $: 
It equals 1 if the spaces are orthogonal and 0 if one is a subspace of the other. 
Similarly, we compute $ d(\mathbf{B}_i, \mathbf{B}_{vc})$ and $ d(\mathbf{B}_i, \mathbf{B}_{t+vc})$.

As shown in Tab.~\ref{tab:subspace}, due to the modality gap, the text classifier space significantly deviates from the image feature space. 
In contrast, the classifier that compensates for modality gap better aligns with the image space. 
Combining both classifiers further improves coverage and highlighting their complementary roles.

\begin{table}[]
    \centering
    \resizebox{\linewidth}{!}{
    \begin{tabular}{cccc}
    \toprule
            &CIFAR-100&ImageNet-R&ImageNet100  \\
    \midrule
    $d(\mathbf{B}_i, \mathbf{B}_t)$& 0.7732&0.7160&0.7664\\
    $d(\mathbf{B}_i, \mathbf{B}_{vc})$&0.6076&0.5843&0.5414\\
    $d(\mathbf{B}_i, \mathbf{B}_{t+vc})$&\textbf{0.4917}&\textbf{0.3284}&\textbf{0.4431}\\
    \bottomrule
    
    \end{tabular}
    }
    \caption{Difference measurement of image subspace and subspace of different classifiers on different datasets.}
    \label{tab:subspace}
\end{table}

\minisection{Analyzing the impact of our method on the modality gap.}
We analyze the impact of our method on the modality gap. Tab.~\ref{tab:keep_gap_ana} presents the mean cosine similarity of the positive and negative for the final task under different experimental settings, along with the last accuracy (Last Acc). 
The `Base' refers to the results of naive fine-tuning. It can be observed that the mean of the negative similarity significantly decreases compared to the original CLIP. This suggests that the modality gap is expanded, leading to a lower last accuracy. 
The `Distill' represents the traditional distillation method, which explicitly limits the output magnitude of the model. It restricts the changes in the modality gap and has some positive effect. However, a complete restriction of the modality gap evidently leads to a decline in the model's learning capability. This leads to suboptimal performance. 
We can see that its performance is close to the original CLIP.
In contrast, our method maintains a relatively stable modality gap, allowing for a moderate increase in the positive similarity and a decrease in the negative similarity. This balance ensures that the model can appropriately learn new knowledge while maintaining pre-trained knowledge, ultimately achieving optimal performance.

\begin{table}[]
    \centering
    \begin{tabular}{ccccc}
    \toprule
         &CLIP  &Base &Distill &Ours  \\
    \midrule
    pos&0.3157   &0.3037&0.3168&0.3251\\
    neg&0.1719   &0.0579&0.1718&0.1520\\
    \midrule
    Last Acc&75.40  &72.74&75.85&\textbf{78.38}\\
    \bottomrule
    \end{tabular}
    \caption{Comparison of cosine similarity and last accuracy (Last Acc) across different experimental settings on ImageNet-100.}
    \label{tab:keep_gap_ana}
\end{table}

%% file: sec/5_conclusion.tex
\section{Conclusion}
This paper investigates the impact of modality gap on the performance of vision-language pre-trained models in class-incremental learning. We find that maintaining a relatively stable modality gap helps preserve pre-trained knowledge and prevents its degradation. Under the condition of a stable modality gap, training a visual space classifier that is not restricted by the modality gap can compensate for some of its negative effects, further enhancing the model's capabilities. The experimental results demonstrate the effectiveness of our approach.

\minisection{Limitations and Future Work.}
Our focus has been on the modality gap, and the model has been fine-tuned simply using LoRA without considering other fine-tuning methods. our current method does not incorporate specially designed loss functions or parameter constraint to mitigate forgetting. Future work will explore the integration of this approach with other continual learning methods and investigate suitable distillation strategies.

\section*{Acknowledgment}
This work was funded by NSFC (NO. 62206135, 62225604), Young Elite Scientists Sponsorship Program by CAST (2023QNRC001), “Science and Technology Yongjiang 2035" key technology breakthrough plan project (2024Z120), Shenzhen Science and Technology Program (JCYJ20240813114237048), and the Fundamental Research Funds for the Central Universities (Nankai Universitiy, 070-63233085). 
Computation was supported by the Supercomputing Center of Nankai University.

%% file: sec/supp.tex
\clearpage
\setcounter{page}{1}
\setcounter{section}{0}
\maketitlesupplementary

\section{Visualization of Modality Gap}
\begin{figure}[h]
    \centering
    \includegraphics[width=0.99\linewidth]{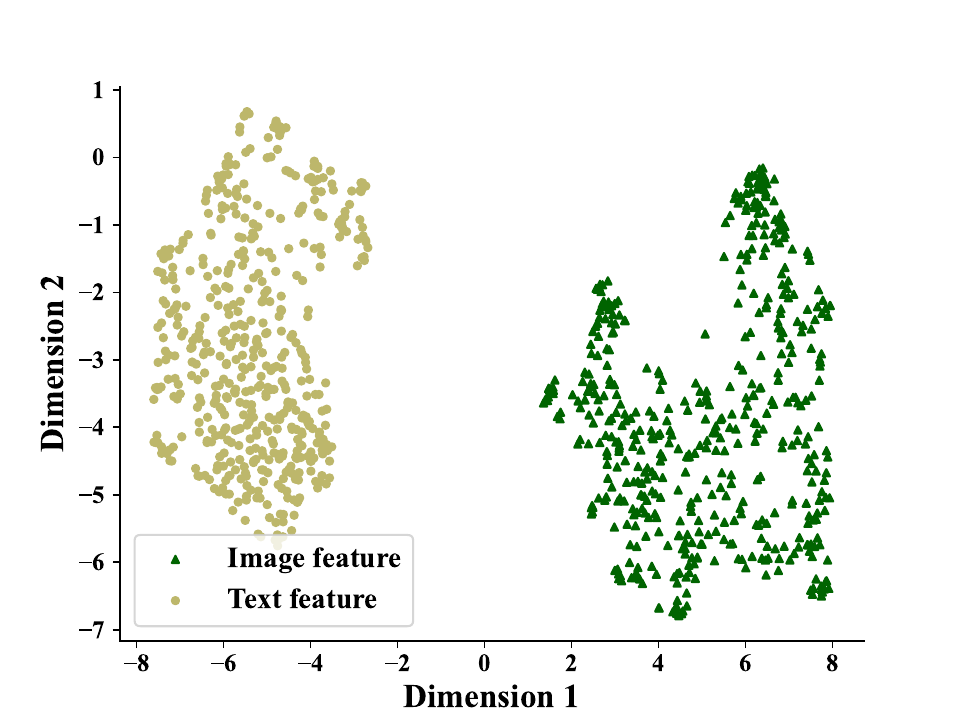}
    \caption{The features of the image and its corresponding caption visualized using UMAP.}
    \label{fig:gap}
\end{figure}

As shown in the Fig.~\ref{fig:gap}, we randomly sampled 512 image-caption pairs from LAION-400M~\cite{schuhmann2021laion}, extracted features using CLIP, and applied UMAP~\cite{mcinnes2018umap} for dimensionality reduction. The results reveal a clear clustering of features within the same modality, while features from different modalities maintain a certain distance. This phenomenon reflects the modality gap.

\section{Proof of Image-Space Classifier and Modality Gap Constraints}

\subsection{Existence of an Optimal Classifier within Image Feature Space}

Let the CLIP image feature matrix be $ \mathbf{X} \in \mathbb{R}^{d \times n} $. Consider a classifier $ \mathbf{W}_* \in \mathbb{R}^{d \times C} $ that achieves minimal cross-entropy loss under ideal conditions. We show that there always exists an equivalent classifier $ \mathbf{W}_{\text{opt}} $ that is entirely contained within the span of $ \mathbf{X} $, i.e., $\mathbf{W}_{\text{opt}} \in  \text{span}(\mathbf{X})$.  

Since any classifier $ \mathbf{W}_* $ can be decomposed as:
\begin{equation}
    \mathbf{W}_* = \mathbf{W}_{\parallel} + \mathbf{W}_{\perp},
\end{equation}
where $ \mathbf{W}_{\parallel} \in \text{span}(\mathbf{X}) $ and $ \mathbf{W}_{\perp} \perp \text{span}(\mathbf{X}) $.
The input feature $\mathbf{x}_i \in \mathbf{X}$ and its label $y$ contribute to the loss function composed of softmax and cross-entropy as follows:
\begin{equation}
\mathcal{L}_{ce} = -\sum_i^n\log \frac{\exp (\mathbf{w}_y^\top\mathbf{x}_i)}{\sum_{j}^C \exp (\mathbf{w}_j^\top\mathbf{x}_i)}
\end{equation}
Obviously, $ \mathbf{W}_{\perp} $ does not relate to the loss.
Thus, an equivalent classifier achieving the same loss can be expressed as:
\begin{equation}
    \mathbf{W}_{\text{opt}}^\top \mathbf{X} = \mathbf{W}_{\parallel}^\top \mathbf{X} = \mathbf{W}_*^\top \mathbf{X}.
\end{equation}
This establishes the existence of an optimal classifier contained within the image feature space.

\subsection{Restriction Imposed by Modality Gap on Text Classifiers}

Using Singular Value Decomposition (SVD), we express $ \mathbf{W}_{\text{opt}} $ as:
\begin{equation}
    \mathbf{W}_{\text{opt}} = \mathbf{U} \mathbf{\Sigma} \mathbf{V}^{\top},
\end{equation}
where $ \mathbf{U} \in \mathbb{R}^{d \times r'} $ represents an orthonormal basis for a subspace of $ \text{span}(\mathbf{X}) $. The text feature matrix $ \mathbf{T} $ can be decomposed into:
\begin{equation}
    \mathbf{T} = \mathbf{T}_{\parallel} + \mathbf{T}_{\perp},
\end{equation}
where $ \mathbf{T}_{\parallel} = \mathbf{U}_r\mathbf{A} $ (with rank $ r \leq r' $) lies within the image feature subspace, and $ \mathbf{T}_{\perp} $ is its orthogonal complement.

Since $ \mathbf{T}_{\perp} $ does not contribute to classification, the optimal alignment is obtained by solving:
\begin{equation}
    \min_{\mathbf{A}} \|\mathbf{U}_r\mathbf{A} - \mathbf{W}_{\text{opt}}\|_F^2.
\end{equation}
The optimal solution is given by:
\begin{equation}
    \mathbf{A}^* = \mathbf{U}_r^{\top} \mathbf{W}_{\text{opt}}.
\end{equation}
Substituting this back, the misalignment error is lower-bounded by:
\begin{equation}
    \|\mathbf{T}_{\parallel} - \mathbf{W}_{\text{opt}}\|_F^2 \geq  \|\mathbf{U}_r\mathbf{U}_r^{\top} \mathbf{W}_{\text{opt}} - \mathbf{W}_{\text{opt}}\|_F^2  
    = \sum\nolimits_{i=r+1}^{r'} s_i^2,
\end{equation}
where $ s_i $ are the singular values of $ \mathbf{W}_{\text{opt}} $.

This result implies that perfect alignment is achievable only when the text feature subspace has sufficient rank to fully capture $ \mathbf{W}_{\text{opt}} $, i.e., when $ r = r' $. However, due to the modality gap, the effective rank of the text classifier subspace is often lower ($ r < r' $), leading to an inherent limitation in classification performance.

\section{Implementation Details of Image Space and Classifier Space Analyzing}
To analyze these relationships, we first apply SVD to the image feature matrix and extract its corresponding basis vectors, denoted as $\mathbf{B}_i \in \mathbb{R}^{d \times r}$. Given the large number of image features, we retain only the basis vectors that capture 95\% of the total energy, reducing noise while preserving essential information.  
For the text feature classifier and the visual space classifier, we employ QR decomposition to obtain their respective basis vectors, denoted as $\mathbf{B}_t$ and $\mathbf{B}_{vc}$. Furthermore, we compute the basis vectors of the combined space spanned by both classifiers, represented as $\mathbf{B}_{t+vc}$.

\section{More experiments}

\subsection{Compatibility with replay methods.}
Our method does not require rehearsal samples; however, it is still compatible with them. We tested our method with simple random sampling of rehearsal data, keeping the total number at 2000. The experimental results, shown in Table~\ref{tab:withreplay}, demonstrate that our method benefits from the replay data and is compatible with it.

\begin{table}[]
    \centering
    \begin{tabular}{ccccc}
    \toprule
        &\multicolumn{2}{c}{CIFAR100}&\multicolumn{2}{c}{ImageNet100}\\
         &Avg&Last&Avg&Last  \\
    \midrule
    ours w/o replay&86.79 &80.40    &87.31&78.38\\
         ours w/ replay&88.48 &82.58    &88.50&80.74\\
    \bottomrule
    \end{tabular}
    \caption{Experimental results with replay data on our method}
    \label{tab:withreplay}
\end{table}

\subsection{Experiments of CLIP ViT-L/14 backbone}
We evaluate the effectiveness of our method on another CLIP model. For this, we replace the backbone of all methods with OpenAI's stronger ViT-L/14 model. The experimental results, shown in Table~\ref{tab:backbone_large}, demonstrate that our method still outperforms the others.

\begin{table}[]
    \centering
    \begin{tabular}{ccccc}
    \toprule
    \multirow{2}{*}{Method}&\multicolumn{2}{c}{CIFAR100}&\multicolumn{2}{c}{ImageNet-R}\\
    &Avg&Last&Avg&Last\\
    \midrule
    PROOF           &89.87&83.59    &91.25&87.33  \\
    CLAP            &87.94&84.86    &92.12&88.63\\
    SLCA            &90.12&84.62    &89.99&86.83\\ 
    RAPF            &90.25&85.29    &91.96&88.32  \\
    L2P++           &85.68&77.86    &90.49&86.73\\ 
    DualPrompt      &86.63&79.12    &90.66&87.14 \\
    CODA            &85.82&78.67    &89.11&84.56\\ 
    Continual-CLIP  &80.48&73.46    &86.99&83.05  \\
    Aper-Adapter    &80.21&71.95    &89.17&85.4\\ 
    MOE4CL          &90.98&85.83    &93.27&90.42\\
    CLAP*           &74.41&71.57    &91.10&87.55\\ 
    MagMax          &90.16&86.06    &93.22&89.55\\
    \midrule
    ours            &91.78&87.03    &93.66&91.08\\
    \bottomrule
    \end{tabular}
    \caption{Experimental Results with ViT-L/14 Backbone Model}
    \label{tab:backbone_large}
\end{table}

\subsection{Hyperparameter selection}
We use the same hyperparameters for all datasets. Below are the experiments for hyperparameter selection on a single dataset.

\paragraph{Rank of LoRA}
As shown in Table~\ref{tab:lora_rank}, our method is insensitive to the rank of LoRA. We choose a rank of 8 for experiments across all datasets.

\begin{table}[]
    \centering
    \begin{tabular}{ccccc}
    \toprule
         Rank&4&8&16&32  \\
    \midrule
         Last&78.02&78.38& 78.32&78.26\\
    \bottomrule
    \end{tabular}
    \caption{Experiments with different ranks of LoRA on ImageNet-100}
    \label{tab:lora_rank}
\end{table}

\paragraph{Output the ensemble weights $\boldsymbol{\beta}$}
As shown in Table~\ref{tab:beta}, as we increase the weight $\boldsymbol{\beta}$ assigned to the visual-space classifier’s output, the overall performance first improves and then declines. This suggests that higher-confidence predictions from the visual-space classifier effectively compensate for the shortcomings of the text classifier, while lower-confidence predictions have minimal impact on the overall results. However, when $\boldsymbol{\beta}$ becomes too large, even low-confidence predictions from the visual classifier can significantly influence the text classifier’s output, allowing incorrect predictions with low scores to dominate. Therefore, an appropriately chosen weight enables better complementarity between the two classifiers.
\begin{table}[]
    \centering
    \begin{tabular}{cccccc}
    \toprule
         $\boldsymbol{\beta}$&1 &2  &4  &6  &8  \\
    \midrule
         Last&77.58 &77.92  &78.38  &77.98  &77.32\\
    \bottomrule
    \end{tabular}
    \caption{Experiments with different $\boldsymbol{\beta}$ on ImageNet-100}
    \label{tab:beta}
\end{table}

\paragraph{Effect of $\boldsymbol{\alpha}$}
As shown in the Tab.~\ref{tab:alpha}, when $\boldsymbol{\alpha}$ is too small, it excessively constrains model training, preventing it from fully learning the new task and limiting performance. Conversely, when $\boldsymbol{\alpha}$ is too large, the pre-trained knowledge is disrupted, leading to a gradual performance decline. We select 10\% as our $\boldsymbol{\alpha}$.

\begin{table}[]
    \centering
    \begin{tabular}{ccccc}
    \toprule
         $\boldsymbol{\alpha}$&5\%& 10\%&20\%&30\% \\
    \midrule
         Last&77.88& 78.38 &77.34&76.9\\
    \bottomrule
    \end{tabular}
    \caption{Experiments with different $\boldsymbol{\alpha}$ on ImageNet-100}
    \label{tab:alpha}
\end{table}

\subsection{Zero-shot Capability}
As shown in Tab.~\ref{tab:ImageNet-C}, we run zero-shot tests on four ImageNet-C~\cite{hendrycks2019imagenet-c} corruptions (severity 3) after continual learning from CIFAR-100.
\textbf{Defocus blur}: Due to CIFAR-100's low resolution, all methods slightly improved over the original CLIP; our method performed best.
\textbf{Other corruptions}: Other methods showed reduced robustness, while ours maintained CLIP performance with minimal drop and slight gains on Contrast.
This suggests our approach maintains CLIP's generalization better than other methods.

\begin{table}[]
\centering
    \resizebox{0.49\textwidth}{!}{
        \begin{tabular}{lcccc}
        \toprule
                &Defocus    &Contrast   &Frost  &Gaussian\\
        \midrule
        CLIP   &41.95      &55.07      &38.23  &43.20\\
        \midrule
         MagMax &43.51      &52.10      &36.75  &41.43\\
         MOE4CL &44.24      &54.82      &34.75  &36.09\\
         Ours   &44.65      &56.02      &37.71  &42.47\\
        \bottomrule
        \end{tabular}
    }
        \caption{Zero-shot performance on ImageNet-C after class-incremental learning on CIFAR-100.}
    \label{tab:ImageNet-C}
\end{table}

\subsection{Few-shot Capability}
As shown in Tab.~\ref{tab:fewshot}, we report last accuracy in continual learning on CIFAR-100 10 task under limited data settings.
Our method shows greater advantage under limited data conditions.

\begin{table}[]
\centering
        \begin{tabular}{lcccc}
        \toprule
                &100-shot        &50-shot         &25-shot         &5-shot  \\
        \midrule
         MagMax &75.82      &75.02      &72.53      &67.66  \\
         MOE4CL &75.52      &75.40       &74.98      &68.54  \\
         Ours   &78.30      &78.04      &77.04      &75.10  \\
        \bottomrule
        \end{tabular}
        \caption{Last accuracy on CIFAR-100 (10-task) in few-shot settings, showing consistent gains of our method under limited data.}
    \label{tab:fewshot}
\end{table}

\subsection{Additional task settings}

We further evaluate our method under different task settings. As shown in Tab.~\ref{tab:multitasks}, while our performance on CIFAR-100 (5-task) is lower than the best baseline, our method achieves the highest accuracy under the more challenging 20-task setting.

\begin{table}[]
\centering
        \begin{tabular}{lcccc}
        \toprule
                &\multicolumn{2}{c}{CIFAR-100}         &\multicolumn{2}{c}{ImageNet-R}  \\
                &5task      &20task     &5task      &20task\\
        \midrule
         MagMax &82.07      &76.84      &82.75      &80.18  \\
         MOE4CL &78.96      &76.20       &81.37      &79.58  \\
         LGVLM\cite{zhang2024LGVLM}  &83.84      &77.26      &82.46      &79.32  \\
         Ours   &81.47      &79.31      &83.13      &82.12  \\
        \bottomrule
        \end{tabular}
        \caption{Last accuracy under different task settings on CIFAR-100 and ImageNet-R.}
    \label{tab:multitasks}
\end{table}

\subsection{Study on Auxiliary Strategies}
To explore potential performance enhancements, we test two auxiliary strategies: (1) \textbf{Exponential Moving Average (EMA)} and (2) \textbf{Per-task Epoch Estimation}.
As shown in Tab.~\ref{tab:aux_strate}, both bring minor improvements but will increase computational cost. In particular, per-task epoch estimation doubles the number of forward passes. To maintain efficiency, we adopt our method without these additions.

\begin{table}[]
\centering
\begin{tabular}{lccc}
\toprule
& Ours & + EMA & + Epoch Est. \\
\midrule
Avg & 87.58 & 87.73 & 87.77 \\
Last & 82.67 & 82.92 & 82.82 \\
\bottomrule

\end{tabular}
\caption{Study of auxiliary strategies on ImageNet-R (10-task).}
\label{tab:aux_strate}
\end{table}

\section{Algorithm Pseudocode}
The overall pipeline of training is shown in the pseudocode:

\begin{algorithm}[]
\caption{Training algorithm}
\label{alg:train}
\begin{algorithmic}[1] %
\State \textbf{Input:} $\mathbf{D} = \{ \mathbf{X}_1, \mathbf{X}_2 ,\dots, \mathbf{X}_t \}$  \Comment{Training data in all tasks}  
\State \textbf{Input:} $f^0_{\text{clip}}(\cdot)$ \Comment{Original CLIP} 

\State \textbf{require:} $f^T_{clip}(\cdot)$ \Comment{CLIP after fine-tuning on $T$ tasks} 
\State \textbf{require:} $\mathbf{W}_v$  \Comment{cosine classifier in visual space} 

\State \textbf{Initialize:} $e = 0$     \Comment{fine-tuning period} 

\For{$t = 1 \text{ to } T$}
    \State $f^{t,0}_{\text{clip}}(\cdot) = f^{t-1}_{\text{clip}}(\cdot)$    \Comment{init model in current task}
    \If{$t = 1$}        \Comment{calculate the fine-tuning period} 
        \State $neg^0 = \textit{Eq.2}(f^{1,0}_{\text{clip}}(\cdot), \mathbf{X}_1)$
        \State $f^{1,1}_{\text{clip}}(\cdot) = \textit{FINETUNE}(f^{1,0}_{\text{clip}}(\cdot), \mathbf{X}_1)$
        \State $neg^{1} = \textit{Eq.2}(f^{1,1}_{\text{clip}}(\cdot),\mathbf{ X_1})$     
        \While{$\textit{Eq.3}(neg^0, neg^{e+1}) < \alpha$}
            \State $e += 1$
            \State $f^{1,e+1}_{\text{clip}}(\cdot) = \textit{FINETUNE}(f^{1,e}_{\text{clip}}(\cdot), \mathbf{X}_1)$
            \State $neg^{e+1} = \textit{Eq.2}(f^{1,e+1}_{\text{clip}},\mathbf{ X_1})$            
        \EndWhile
        $e = max(1,e)$
    \Else
        \For{$i = 1 \text{ to } e$}
            \State $f^{t,i}_{\text{clip}}(\cdot) = \textit{FINETUNE}(f^{t,i-1}_{\text{clip}}(\cdot), \mathbf{X}_t)$
        \EndFor
    \EndIf
    \State $f^{t}_{\text{clip}}(\cdot) = f^{t,e}_{\text{clip}}(\cdot)$
    \State \textbf{Initialize:} $\mathbf{W}_v^t$   \Comment{initalize $\mathbf{W}_v^t$ based on class prototypes} 
    \State $W_v^t = \textit{TRAIN}(\mathbf{W}_v^t, f^{t}_{\text{clip}}(\mathbf{X_t}))$
\EndFor

\State \textbf{return} $f^{T}_{\text{clip}}(\cdot), \mathbf{W}_v^T$ 

\end{algorithmic}
\end{algorithm}